# ClickSAM: Fine-tuning Segment Anything Model using click prompts for ultrasound image segmentation

Aimee Guo[a], Grace Fei[b], Hemanth Pasupuleti[c], Jing Wang[d]

[a]The Hockaday School, 11600 Welch Rd, Dallas, TX 75229; [b]Liberty High School, 15250 Rolater Rd, Frisco, TX 75035; [c]The University of Texas at Dallas, 800 W Campbell Rd, Richardson, TX 75080; [d]The University of Texas Southwestern Medical Center at Dallas, 5323 Harry Hines Blvd, Dallas, TX 75390

## Abstract

The newly released Segment Anything Model (SAM) is a popular tool used in image processing due to its superior segmentation accuracy, variety of input prompts, training capabilities, and efficient model design. However, its current model is trained on a diverse dataset not tailored to medical images, particularly ultrasound images. Ultrasound images tend to have a lot of noise, making it difficult to segment out important structures. In this project, we developed *ClickSAM*, which fine-tunes the Segment Anything Model using click prompts for ultrasound images. ClickSAM has two stages of training: the first stage is trained on single-click prompts centered in the ground-truth contours, and the second stage focuses on improving the model performance through additional positive and negative click prompts. By comparing the first stage's predictions to the ground-truth masks, true positive, false positive, and false negative segments are calculated. Positive clicks are generated using the true positive and false negative segments, and negative clicks are generated using the false positive segments. The Centroidal Voronoi Tessellation algorithm is then employed to collect positive and negative click prompts in each segment that are used to enhance the model performance during the second stage of training. With click-train methods, ClickSAM exhibits superior performance compared to other existing models for ultrasound image segmentation.

## 1. Introduction

Segmentation is a crucial step in medical image processing, which involves the identification of distinct components within an image. Accurate segmentation of regions of interest (ROI) such as tumors and organs in medical imaging plays a critical role in both diagnosis (e.g., quantitative imaging analysis) and treatment planning (e.g., in radiotherapy and surgical planning). While manual segmentation is still widely used in clinics, it is very laborintensive, time-consuming, and requires a high level of expertise. More efficient and accurate segmentation methods are highly desired.

Automated segmentation models have been under rapid development in recent years, enabled by artificial intelligence and machine learning. As a large-scale foundation model, the newly released Segment Anything Model (SAM) [1] is an innovative image segmentation method that makes segmentation easier and more efficient with its exceptional accuracy. Utilizing SAM, only a few clicks are used to segment out ROI without requiring task-specific annotation, training, and modeling. SAM's input prompts reflect how clinicians analyze images: clicking on the screen to select parts of the image or utilizing bounding boxes to crop parts out. Though SAM has shown great promise in medical image segmentation, there are several issues when it is applied to ultrasound image segmentation. Ultrasound images contain a lot of noise that make it difficult for the Segment Anything Model to accurately segment out organs, or more importantly, tumors and cancerous regions. In addition, SAM is trained on a large diverse dataset including images not specific to ultrasound, making the final segmentation less accurate than expected. This makes fine-tuning especially important, and current models such as MedSAM [2] address this problem by fine-tuning SAM with bounding box prompts on various types of medical images. Inspired by Segmentation Click Train [3], we propose *ClickSAM*, which focuses on fine-tuning SAM by utilizing click prompts to improve ultrasound image segmentation accuracy. Segmentation Click Train [3] utilizes SegFormer [4] in training, while ClickSAM utilizes SAM to take advantage of its high segmentation accuracy. By utilizing click-train methods to finetune SAM, ClickSAM exhibits superior performance compared to these existing segmentation models [2, 3] when tested on ultrasound images.

## 2. Methods

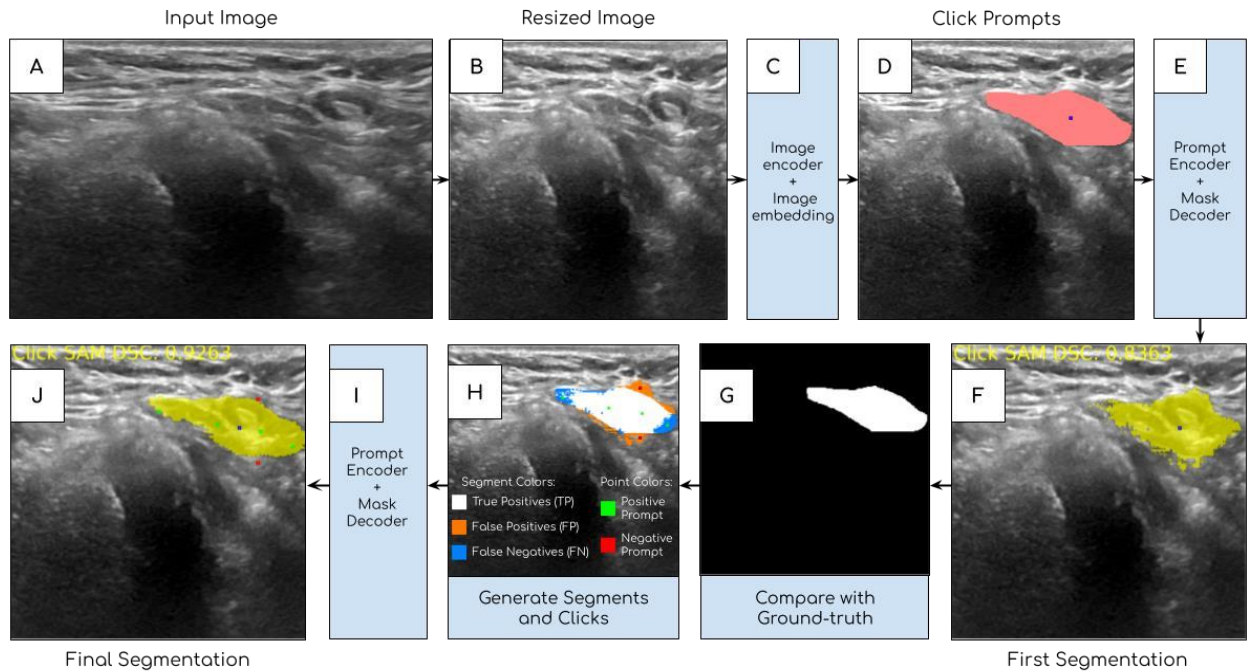

**Fig. 1.** ClickSAM has two stages of training: the first stage is trained on single-click prompts shown as blue dots centered in the ground-truth segment (top row), and the second stage focuses on improving the model through additional positive and negative click prompts shown as green and red dots respectively (bottom row). Both training and testing follow the two-stage process.

The proposed ClickSAM is built on the Segment Anything Model (SAM) from Meta AI because of its easy manipulation and ability to be incorporated into other systems. SAM can accept multiple input prompts such as points, boxes, and masks, proving it a preferred platform to develop ClickSAM. In ClickSAM, an image is first resized to 256×256 pixels (Fig. 1 A and 1 B). Utilizing the SAM network architecture, the image encoder produces an image embedding (Fig. 1 C) where images are stored as a vector for further processing. Single-point prompts shown as blue dots are calculated as the centers of mass of the ground-truth masks and are fed into the prompt encoder (Fig. 1 D). They are then provided with the image embedding to the mask decoder, which generates the first predictions used for further improvement (Fig. 1 E and F). These initial predictions are then compared with corresponding groundtruth masks to calculate connected components (i.e., regions that are connected) with true positives (white), false positives (orange), and false negatives (blue). Connected components (shown in Fig. 1 H) are then used to generate clicks, also known as point prompts. The number of clicks per region will vary depending on the size of the region; each region will contain clicks unless the region is too small. Clicks are generated with the Centroidal Voronoi Tessellation algorithm [5] to ensure clicks are evenly spaced in the segments. False negatives and true positive clicks are considered positive prompts (green dots) because those segments are in the ground-truth mask and should be in the prediction. On the other hand, false positive clicks are considered negative prompts (red dots) because they should not be in the prediction. Different types of clicks are illustrated in Fig. 1 H where the green clicks represent both true positive and false negative segments, while the red clicks only contain false positive segments. The model is then trained again with SAM's network architecture using the positive and negative click prompts to improve the accuracy of the model (Fig. 1 I) to generate the final segmentation (Fig. 1 J). The positive and negative prompts give the model additional information about where to segment; whether the model needs to segment out a certain area that it didn't include previously, or if the model needs to get rid of a certain area previously included. In this project, dice and cross-entropy loss were used as the training loss, like MedSAM [2].

## 3. Experiments and Results

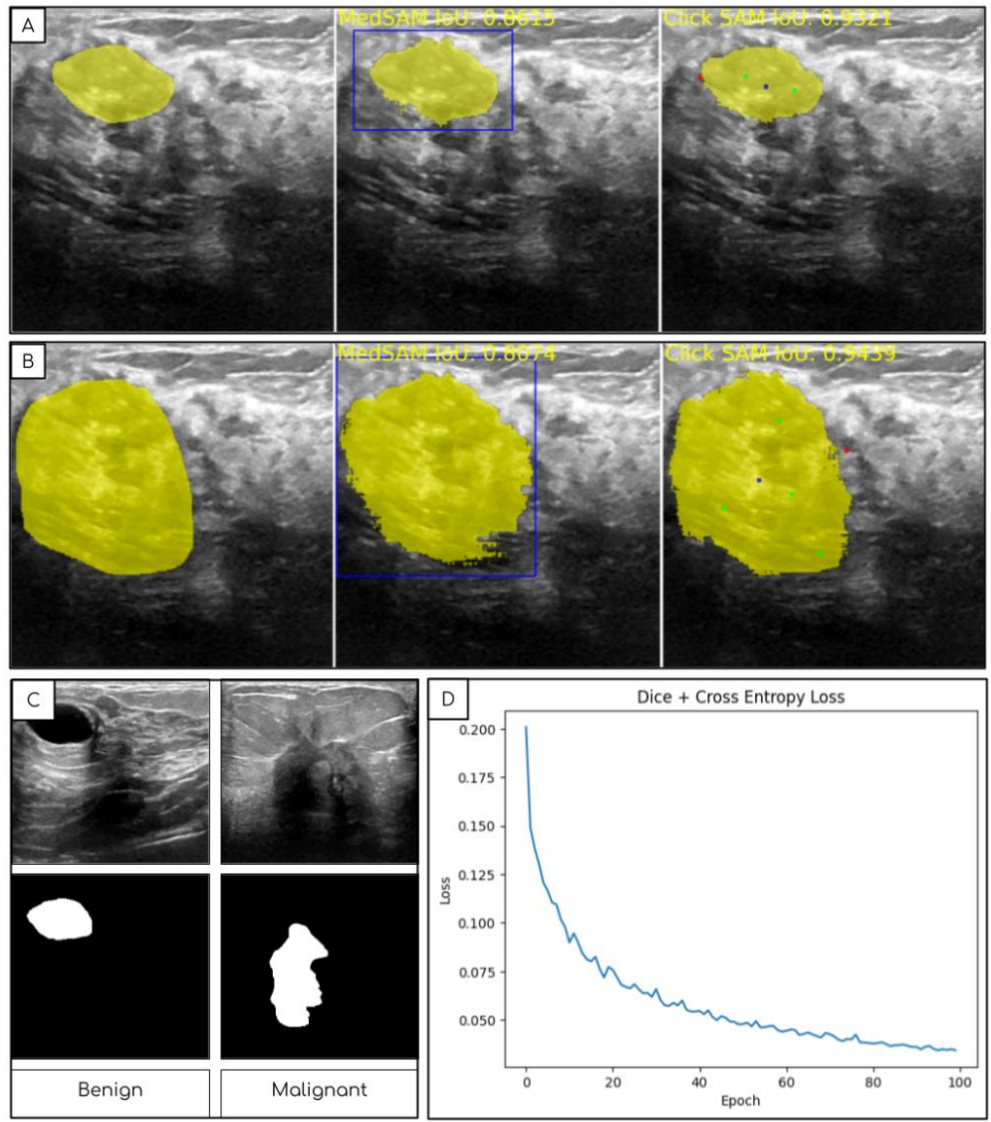

**Fig. 2.** ClickSAM's experimental results. **(A & B)** ClickSAM's IoU is significantly higher than MedSAM's IoU as shown in the example image. **(C)** Sample images used in this project are from the Dataset of Breast Ultrasound Image (BUSI) [6]. The two types of data employed in training are benign and malignant images. **(D)** The Loss v. Epoch graph shows the improvement of the model during the training epochs.

ClickSAM was coded in Python on Visual Studio Code and ran on Jupyter Notebook on an Ubuntu computer with a NVIDIA GeForce RTX 3080 GPU, 32 GB memory, and 11th Gen Intel Core i7-11700K CPU. The proposed ClickSAM model was trained on the Dataset of Breast Ultrasound Images (BUSI) [6], which includes 647 images that are categorized as 2 different classes: benign and malignant (Fig. 2 C). ClickSAM splits the dataset up by randomly categorizing 80% for training and 20% for testing, just like MedSAM [2] does. The dice and cross-entropy loss decreased significantly over the model training process (Fig. 2 D). After training, Intersection over Union (IoU) was calculated to measure the model's performance. IoU measures the overlap between the groundtruth's bounding box with the prediction's bounding box. ClickSAM achieved a mean IoU of 0.916, while MedSAM achieved a mean IoU of 0.863 and Segmentation Click Train [3] achieved a mean IoU of 0.707. It can be observed that the additional prompts (i.e., clicks) provided during the second training improved the quality of the segmentation drastically. As seem in Fig. 2 A, MedSAM's IoU is 0.8615 while ClickSAM's IoU is 0.9321, and in Fig. 2 B, MedSAM's IoU is 0.8074 while ClickSAM's IoU is 0.9439. In addition, ClickSAM consistently segments out regions in around half a second, showing how quickly and easily the model can perform. These results demonstrate the advantage of using positive and negative click prompts to fine-tune SAM for more accurate lesion segmentation in breast ultrasound imaging.

## 4. Breakthrough Work to be Presented

This paper presents ClickSAM, a fine-tuned Segment Anything Model with both positive and negative click prompts. Training with click prompts in ClickSAM leads to a more accurate model compared to other approaches such as MedSAM [2], which uses bounding box prompts only, as measured by IoU. Integration of click prompts into the fine-tuning process empowers the model to generate more precise coordinates compared to relying solely on bounding boxes. Unlike bounding boxes that define only rectangular regions, click prompts offer a more intricate definition of desired areas, reducing the inclusion of unwanted regions. In SAM's framework, positive and negative bounding boxes are not simultaneously permissible during fine-tuning, and each segment can be assigned only one bounding box, diminishing the effectiveness of bounding box prompts in this context. Furthermore, it's important to note that bounding boxes are constrained by their axis-aligned nature, which renders them less compatible with segmentations that deviate from axis alignment. For instance, if the segmentation assumes a diagonal rectangle shape, a bounding box will encompass numerous undesired regions, thereby diminishing its utility in contrast to the precision offered by click prompts. The adoption of ClickSAM and the incorporation of click prompts demonstrate a superior approach in achieving more accurate and contextually relevant segmentations, overcoming limitations associated with traditional bounding box methods.

## 5. Conclusion and Future Work

ClickSAM uses both positive and negative prompts to fine-tune SAM in a two-stage training process, demonstrating its ability to accurately segment out tumors and cancerous regions in breast ultrasound images, aiding in quantitative imaging analysis and diagnosis. In the future, ClickSAM can incorporate the idea of PseudoClick [7], where no user-supplied prompt is necessary for testing, as the model utilizes an error decoder to predict false positives and false negative regions and generate prompts automatically. ClickSAM can be expanded to other imaging modalities, such as MRI or CT scans and used for other diseases as a medical imaging segmentation model.